\documentclass{article}
\usepackage{graphicx} 
\usepackage{gubbins/iclr2024_conference,times}
\usepackage{url}
\usepackage{float}
\usepackage{amssymb}
\usepackage{amsmath}
\usepackage{listings}
\usepackage{wrapfig}
\usepackage{graphicx}
\usepackage{subcaption}
\usepackage{float}
\usepackage{hyperref}
\usepackage{tcolorbox}
\title{Multi-objective Reinforcement learning from AI Feedback}
\author{Marcus Williams \\ \texttt{marcusjw@zoho.com}}

\date{June 2024}

\begin{document}

\maketitle
\begin{abstract}
This paper presents Multi-Objective Reinforcement Learning from AI Feedback (MORLAIF), a novel approach to improving the alignment and performance of language models trained using reinforcement learning from AI feedback (RLAIF). In contrast to standard approaches that train a single preference model to represent all human preferences, MORLAIF decomposes this task into multiple simpler principles, such as toxicity, factuality, and sycophancy.  Separate preference models are trained for each principle using feedback from GPT-3.5-Turbo. These preference model scores are then combined using different scalarization functions to provide a reward signal for Proximal Policy Optimization (PPO) training of the target language model. Our experiments indicate that MORLAIF outperforms the standard RLAIF baselines and that MORLAIF can be used to align larger language models using smaller ones. Surprisingly, the choice of scalarization function does not appear to significantly impact the results. 
\end{abstract}

\section{Introduction}
Recent advancements in large language models (LLMs) have led to remarkable performance across a wide range of natural language tasks (\cite{bommasani2021opportunities,brown2020language}). However, ensuring that these models behave in alignment with human values and preferences remains a significant challenge (\cite{kenton2021alignment}). Reinforcement learning from human feedback (RLHF) has emerged as a promising approach to address this issue by training models to optimize for human-specified reward functions (\cite{christiano2017deep,stiennon2020learning}), but many issues with RLHF have been identified, such as the limited ability of humans to evaluate responses and reward hacking of the preference model (\cite{casper2023open}).
In a standard RLHF setup, a preference model is trained on human comparisons of model outputs to represent human preferences. This preference model is then used to provide an RL training signal to update the target language model's behavior. Reinforcement Learning from AI Feedback (RLAIF) outsources the human comparison step to another language model. In \cite{bai2022constitutional}, they train the preference model based on a set of ``constitutional'' principles that the model should adhere to. However, these principles are often quite general, containing many different desiderata, and the preference modeling task of capturing all human preferences in a single model is highly complex.
To address these limitations, we propose Multi-Objective Reinforcement Learning from AI Feedback (MORLAIF). This involves decomposing the preference modeling into multiple distinct principles, such as factuality, toxicity, and conciseness, that are more specific and well-defined. Separate preference models are trained for each of these principles using feedback from strong language models, in our case GPT-3.5-Turbo. During RL training, the outputs from these preference models are combined using a scalarization function to provide the reward signal for optimizing the target language model.
MORLAIF has several appealing properties. First, by breaking down preference modeling into specific principles, the feedback collection and preference model training becomes a more straightforward and well-defined task, which we hypothesize will lead to improved preference model performance and better alignment of the resulting language models. Second, MORLAIF allows for training on a large number of principles without necessarily diluting the reward signal. Third, the multi-objective framing provides a more interpretable system, where we can examine how model outputs score on each principle. Finally, MORLAIF allows for easily adjusting preference model behavior by modifying the scalarization function without expensive retraining of the preference models.
\section{Multi-Objective Reinforcement Learning and scalarization functions}
\label{sec:MORL}
Multi-objective reinforcement learning (MORL) is a framework that enables the learning of policies to simultaneously optimize multiple objectives, which may potentially conflict with each other. MORL has been extensively studied in various domains, particularly in robotics \cite{roijers2013survey} and game-playing \cite{mossalam2016multi}. One of the primary challenges in MORL lies in combining the diverse objectives into a single reward signal that can be used by reinforcement learning algorithms. Several approaches have been employed to address this challenge, including scalarization methods such as weighted sums \cite{van2013}, thresholding \cite{gabor1998}, and Pareto-based methods \cite{van2014mutli}.

For our setup, we have a set of reward functions $R_1, \ldots, R_n$ and wish to maximize:
\begin{equation}
J(\pi) = \mathbb{E}_{\xi \sim \pi}\left[f\left(\sum_{t=0}^{\infty} \gamma^t R_1\left(s_t, a_t, s_{t+1}\right), \ldots, \sum_{t=0}^{\infty} \gamma^t R_n\left(s_t, a_t, s_{t+1}\right)\right)\right]
\end{equation}
where $\pi$ is the policy, $\xi$ is a trajectory sampled from the policy, $\gamma$ is the discount factor, and $f$ is a scalarization function that combines the multiple objectives into a single scalar value. The scalarization function $f$ is used to balance the importance of each objective and provide a single reward signal for our RL algorithm to optimize. In our language modelling case we can drop the $s_{t+1}$ as we have deterministic transitions.

To ensure that the optimization remains tractable with gradient-based methods, we must constrain the scalarization function $f$ to be monotonic in each argument (\cite{roijers2013survey}). A scalarization function $f(u_1, \ldots u_i, \ldots u_n)$ is said to be monotonic in an argument $u_i$ if:

\begin{align*}
& \forall x, y \; (x \geq y \implies f(u_1, \ldots, x, \ldots, u_n) \geq f(u_1, \ldots, y, \ldots, u_n)) \ \lor \\ & \forall x, y \; (x \geq y \implies f(u_1, \ldots, x, \ldots, u_n) \leq f(u_1, \ldots, y, \ldots, u_n))
\end{align*}

In this paper we will test the following scalarization functions:
\begin{itemize}
\item Weighted Linear Combination: Perhaps the simplest MORL objective is a weighted linear combination. Theoretically, a weighted linear combination MOMDP can always be reformulated as a standard MDP with a single reward function (\cite{skalse2023}), thus not gaining anything over just using one reward function. In practice however, using MORL often performs better as it eases the specification of said reward function. Max-average is a special case of a linear combination where all the weights are equal to 1. This scalarization function has the disadvantage that it requires an explicit weighting of our principles, as a starting point we will find weights using a logistic regression on a human preference dataset. Mathematically a weighted linear combination looks like:
\begin{equation}
f(R_1, \ldots, R_n) = \sum_{i=1}^{n} w_i R_i
\end{equation}
\item Worst-Case Optimization: This scalarization function which is also known as minimax, the Rawlsian social welfare function or max-min optimization, aims to find a policy that performs well in the worst-case scenario among all the objectives. Here we seek to maximize the minimum reward across all objectives. Minimax optimization is particularly useful in risk-averse domains where the agent wants to ensure a certain level of performance for all objectives. However, this approach may lead to overly conservative policies and may not fully exploit the potential trade-offs between objectives. Mathematically:
\begin{equation}
f(R_1, \ldots, R_n) = \min_{i} \{R_i\}
\end{equation}
\item Soft max-min: Soft max-min is similar to Worst-Case Optimization except that we optimize the softmin instead of the minimum. In theory this should give us some of the benefits of max-min but lead to less conservative policies. This scalarization introduces an extra hyperparameter, the temperature $T$, which controls how soft the softmin is. This hyperparameter may need some tuning for good results.  $f$ becomes:
\begin{equation}
f(R_1, \ldots, R_n) = \sum_{i=1}^n \text{softmin}(R_i) R_i = \sum_{i=1}^n \frac{e^{-\frac{R_i}{T}}}{\sum_{j=1}^n e^{-\frac{R_j}{T}}} R_i
\end{equation}
\item Uncertainty-Weighted Optimization:
Uncertainty-Weighted Optimization takes into account the uncertainty associated with each objective when making decisions. In this approach, we subtract a term proportional to the variance between the different reward functions. This means that for cases where the reward functions agree more we have a larger weight update than when there is a large difference between the reward functions. This scalarization function introduces the extra hyperparameter $\lambda$ which controls how much to punish uncertainty. \cite{coste2024reward} notes that uncertainty weighted optimization works well for most reasonable choices of $\lambda$. Mathematically our scalarization function is:
\begin{equation}
f(R_1, \ldots, R_n) = \frac{1}{n} \sum_{i=0}^{n} R_i - \lambda \frac{1}{n} \sum_{i=0}^{n} \left( R_i - \frac{1}{n} \sum_{i=0}^{n} R_i \right)^2
\end{equation}
\item Lower Quantile Optimization:
Lower quantile optimization focuses on optimizing a lower quantile of the reward functions. This is somewhat related to worst case optimisation where only the lowest reward matters, here instead the lowest quantile $Q_\alpha$ of rewards matter, where $\alpha$ is the fraction of reward functions we want to include. In this paper we test $\alpha=1/3$. Lower quantile optimization shares similarities with worst-case optimization by prioritizing the objectives with the lowest scores. However, it considers multiple rewards simultaneously, leading to less conservative policies compared to worst-case optimization. Mathematically, the lower quantile optimization scalarization function is expressed as:
\begin{equation}
f(R_1, \ldots, R_n) = Q_\alpha(\{R_i\}_{i=1}^{n})
\end{equation}
\item Max-Median: As the name suggests Max-median optimizes the median reward. It is likely more robust to outliers than max-average but we lose the flexibility of an arbitrary linear weighting.
\begin{equation}
f(R_1, \ldots, R_n)= \text{median}(\{R_i\}_{i=1}^{n})
\end{equation}
\item Bernoulli-Nash: The Bernoulli-Nash social welfare function is a scalarization approach that aggregates reward functions multiplicatively, unlike the additive aggregation used in other methods. It is notable for being the only scalarization function that simultaneously satisfies several theoretically desirable properties, including Pareto Optimality, Symmetry, Scale Invariance, and Independence of Irrelevant Alternatives \cite{kaneko1979nash}. Mathematically this function takes the form: 
\begin{equation}
f(R_1, \ldots, R_n) = \sqrt[n]{\prod_{i=1}^{n} R_i}
\end{equation}
\end{itemize}

\section{Methodology}
\begin{figure}[H]
    \centering
    \includegraphics[scale=0.3]{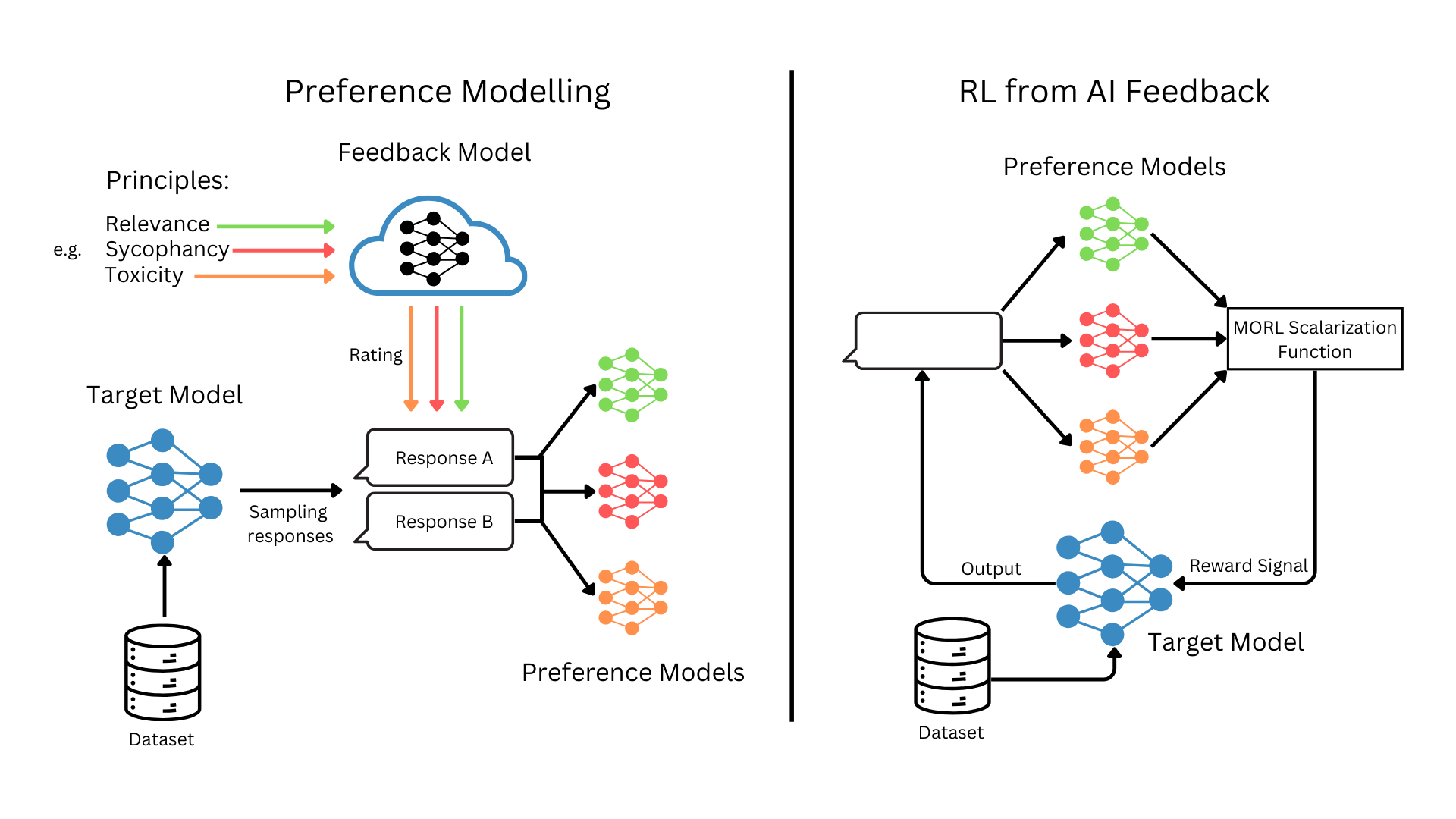}
    \caption{Diagram illustrating the training setup.}
    \label{fig:MORLAIF}
\end{figure}
The methodology consists of two main stages which are illustrated in figure \ref{fig:MORLAIF}, the preference modeling stage and the reinforcement learning from AI feedback stage.
\subsection{Preference Modeling:}
First a SFT model is created by finetuning a base model on the openassistant-guanaco dataset (\cite{dettmers2023qlora}).
The SFT model then generates pairs of responses for a given set of prompts, in this case the prompt portion of Anthropic's HH-RLHF (\cite{bai2022training}).
A feedback model then evaluates which response is better for each individual principle defined in the study.
The ratings from the feedback model are used to train separate preference models for each principle. The preference models can be either full models or Low Rank Adaptations (LoRAs) (\cite{hu2021lora}). 

\subsection{Reinforcement Learning from AI Feedback:}

During the RL loop a MORL scalarization function is used to combine the ratings from each preference model. The different scalarization functions tested are listed in section \ref{sec:MORL} above.
The combined score from the scalarization function serves as a reward signal for training the target model using Proximal Policy Optimization (PPO). We use the implementation of PPO use in \cite{ziegler2020finetuning}. The initial checkpoint used is the same SFT model we trained on openassistant-guanaco.

\subsection{Model details}
The study uses 5 target models, namely GPT-2-medium/large/XL (\cite{radford2019language}), Gemma-2B (\cite{gemmateam2024gemma}) and Llama-7B (\cite{touvron2023llama}). For the preference models both full finetunes and LoRAs were experimented with, but based on results from GPT-2 it was determined that LoRAs perform just as well as full models and consume ~0 extra VRAM compared to single objective RLAIF during the RLAIF phase so these were primarily used. The preference models are typically based on the same model as the one being trained, although we also conduct one experiment where we train Llama-7B using GPT-2-medium as the preference models. GPT-3.5-Turbo was employed as the feedback model to rate response pairs according to each principle. 

\subsection{Evaluation}
We evaluate both the performance of the multi-objective preference models and the performance of the final models.
The PMs are evaluated by measuring their accuracy on the HH-RLHF testset.

We evaluate the performance of the target models by comparing their win rates against a single-objective baseline. We do this in two ways: first, we employ human crowdworkers for two of our most interesting comparisons; second, we utilize GPT-4-Turbo, for all comparisons to provide a comprehensive assessment of the target models' performance.

For the human win-rates we recruited people via the crowdworking platform Amazon MTurk. Workers were asked to have a conversation where two options were presented at each step. These options came from two different models and the human was asked to rate which response was better or if they were equally good. The conversations lasted a maximum of 10 turns or until the human choose to start a new one. The conversations lasted on average 5.2 turns. For more details on how we conducted this and how we reduced ``cheating'' with LLMs see appendix \ref{sec:human}.

For the GPT-4-Turbo win-rates we used the HH-RLHF test set and generated one response for each model and then prompted GPT-4-Turbo to rate which of the two was better. The preferred response was the one with higher token probability. Unlike in the human study we do not give GPT-4-Turbo the option of choosing ``Tie''.

\subsection{Baselines}
For the single-objective RLAIF baseline we essentially copied the procedure in \cite{bai2022constitutional}. We sample principles from their constitution and use the same prompt they used. We also include few-shot examples in the same way they did.

Another baseline which was used was training 12 preference models on the standard RLAIF setup and optimizing our MORL scalarization functions with these PMs. This would eliminate the possibility that the benefit arises from using many PMs, along the lines of \cite{coste2024reward}, rather than from decomposing human preferences.

\subsection{Principles}
The full list of principles used is: helpfulness, ethicality, factuality, toxicity, sycophancy, empathy, relevance, context, bias, understandability, repetitiveness, detail and conciseness. 

Unless overwise stated, we use the full 12 principles. We also conduct an ablation experiment with using fewer principles and see how that reduces performance. These principles were inspired by Anthopic's constitution (\cite{bai2022constitutional}) and the principles used in \cite{go2024compositional}.

\subsection{Code}
The code used in this project can be found at: \url{https://github.com/carolius/Multi-Objective-Reinforcement-Learning-from-AI-Feedback}
\section{Results}
\subsection{Preference Model Accuracy:}
\begin{figure}
\centering
\begin{subfigure}[b]{0.49\textwidth}
\centering
\includegraphics[width=\linewidth]{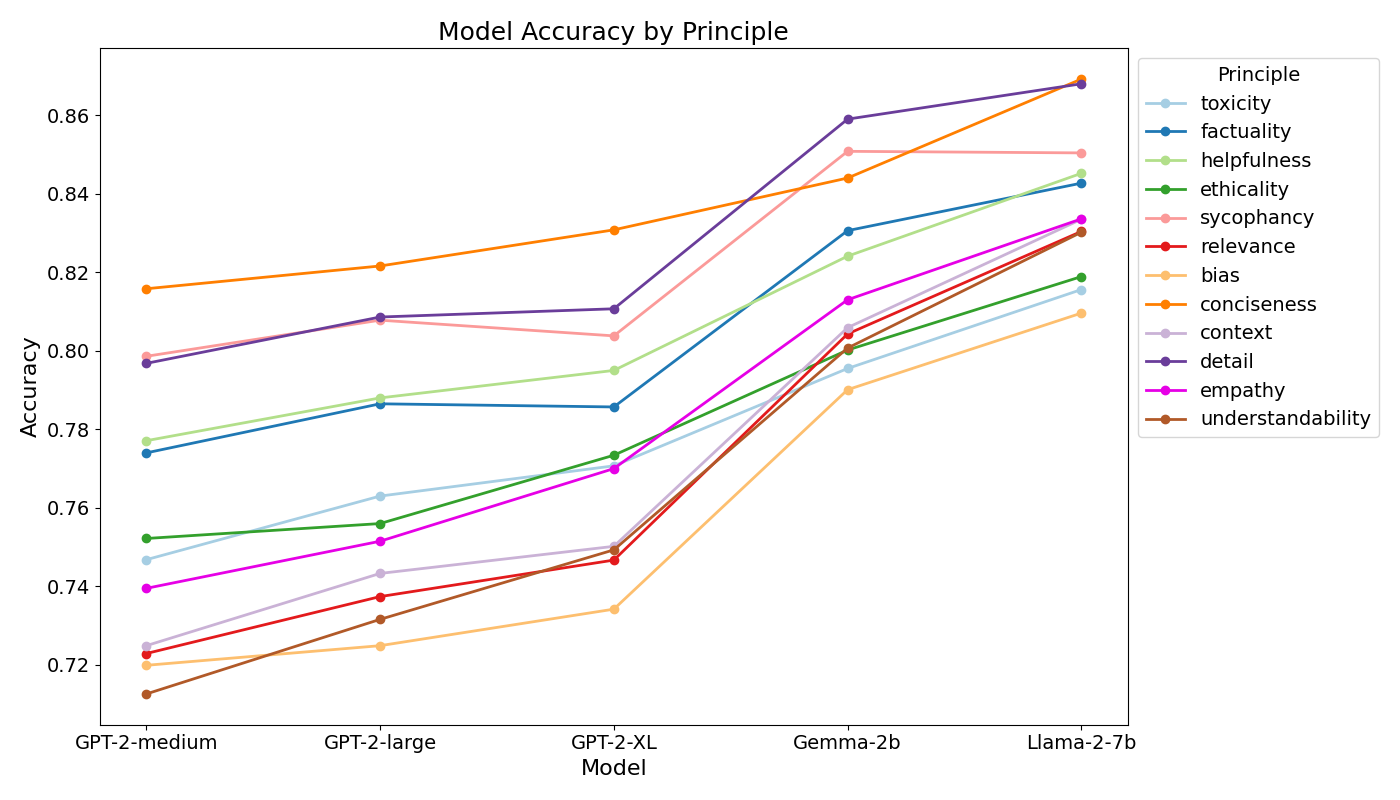}
\caption{}
\label{fig:principle_accuracies}
\end{subfigure}
\hfill
\begin{subfigure}[b]{0.49\textwidth}
\centering
\includegraphics[width=\linewidth]{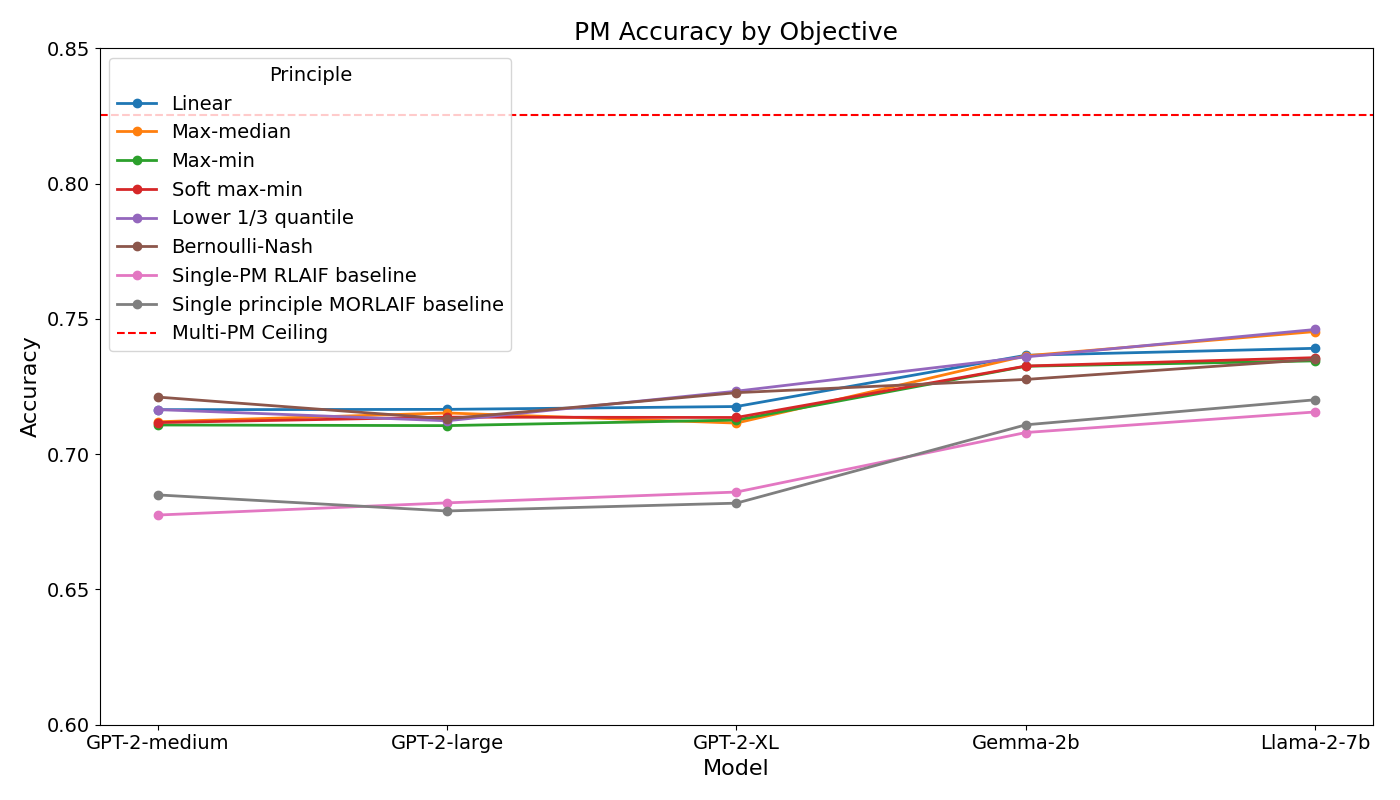}
\caption{}
\label{fig:objective_acc}
\end{subfigure}
\caption{(a) shows the accuracy the different models achieve when trained as a preference model for one of the different principles. (b) shows the accuracy of our Multi-Objective PMs for the different scalarization functions and our single objective baselines.}
\label{fig:accs}
\end{figure}
In figure \ref{fig:principle_accuracies} we can see the accuracy increases with model size as expected. Comparing to figure \ref{fig:objective_acc} we can see that the accuracies for specific principles are generally much higher than for PMs on the true labels. The accuracies also seem to grow faster with increasing model size than in the single principle case. There is quite a large difference between the highest and lowest accuracy principle,  roughly 0.1 for GPT-2-medium.

Figure \ref{fig:objective_acc} shows the accuracy of different MORL scalarization functions. Note that uncertainty weighted scalarization is not included here since the way in which it differs from linear is intended to affect step sizes during training, which is not relevant here. We can see that all MORL objectives perform better than standard single PM RLAIF, but surprisingly they all seem to perform equally well despite being quite different. Based on this data, it is not possible to say which, if any, of the objectives performs the best. \cite{bakker2022finetuning} also note that different social welfare functions seem to behave the same in practice, although their setup is significantly different from ours. We can also note that the accuracy for multi-objective GPT-2-medium is slightly higher than for single-objective Llama-2-7b, despite having roughly 1/20 the number of parameters and being trained on roughly 1/250 the number of tokens. This prompted an experiment to align Llama-2-7b using GPT-2-medium. The figure also includes the performance ceiling of using our 12 principles and represents having PMs that have 100\% accuracy on every principle.
\subsection{Win Rates:}
\begin{figure}[H]
\centering
\begin{subfigure}[b]{0.49\textwidth}
\centering
\includegraphics[width=\linewidth]{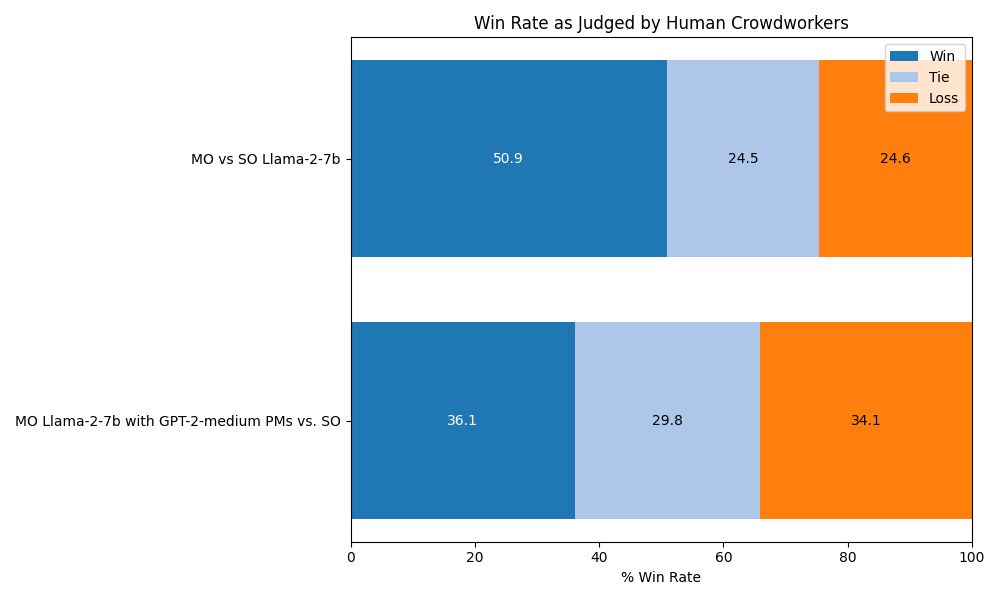}
\caption{}
\label{fig:human_winrate}
\end{subfigure}
\hfill
\begin{subfigure}[b]{0.49\textwidth}
\centering
\includegraphics[width=\linewidth]{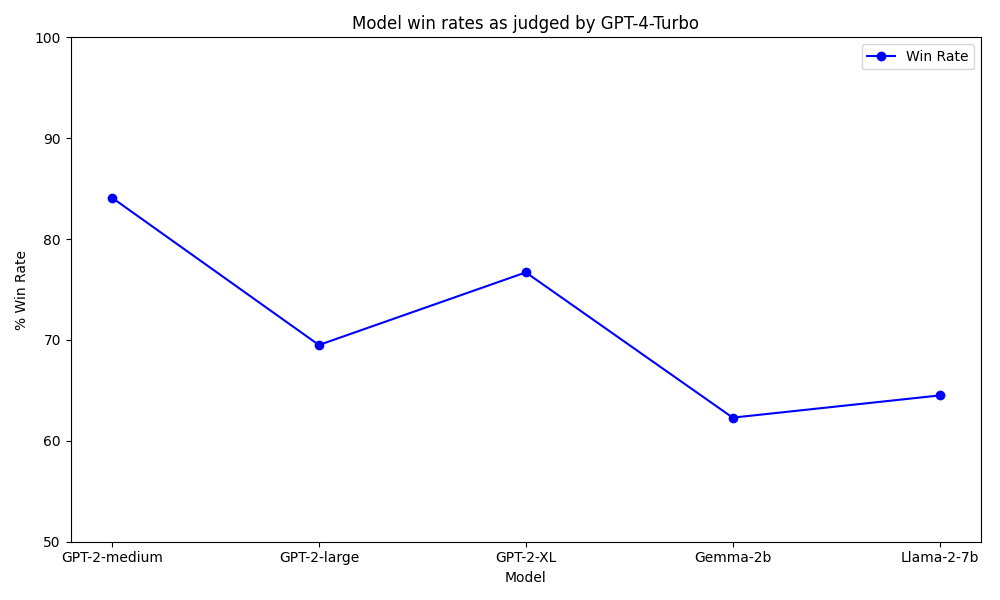}
\caption{}
\label{fig:LLM_winrate}
\end{subfigure}
\caption{(a) shows the human judged win-rate of MORLAIF Llama-2-7b and one trained with GPT-2-medium PMs over the single objective version. (b) shows the GPT-4-Turbo judged win-rate for the different models. }
\label{fig:win_rates}
\end{figure}
Figure \ref{fig:human_winrate} shows the results of the human crowdworker preference experiments. We can see that MORLAIF Llama-2-7b is preferred to Single Objective RLAIF by a large margin, and that a version trained with GPT-2-medium preference models performs the same as the SORLAIF model. As discussed earlier, the fact that using a much weaker multi-objective model results in the same performance as a stronger single-objective model indicates that MORLAIF could be a promising scalable oversight solution.

Figure \ref{fig:LLM_winrate} instead shows the win rates as judged by GPT-4-Turbo. Unlike the human experiments, GPT-4 did not get the option of a tie. We see that the win rate is very high for GPT-2-medium but that it decreases with model size, although it is still significant. We can also note that the GPT-4 rated win rate for Llama-2-7b is approximately the same as the human-rated win rate + tie rate, indicating that GPT-4 is probably a reasonably accurate proxy for human evaluation.
\begin{figure}
\centering
\includegraphics[scale=0.3]{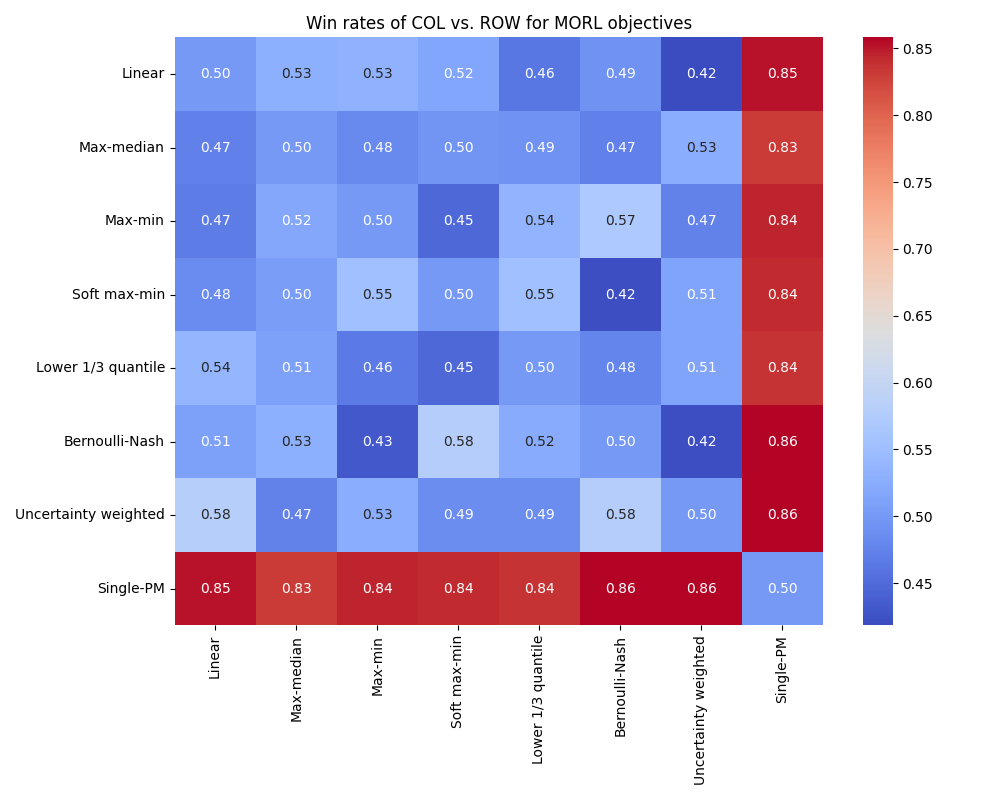}
\caption{Comparison between the win-rates for the different MORL scalarization functions. Note that $e_{i,j} = 1 - e_{j,i}$. }
\label{fig:obj_winrate}
\end{figure}
It could have been the case that while the different MORL objectives score the same in terms of accuracy, they differ in how they affect the final RLAIF model. Figure \ref{fig:obj_winrate} excludes that possibility, at least when using GPT-2-medium according to GPT-4. While again all MORL objectives result in a model much better than using a single PM, the differences between them are indistinguishable from noise.
\subsection{Correlation Between the Principles and Ablation of Principles}
\begin{figure}[H]
\centering
\begin{subfigure}[b]{0.49\textwidth}
\centering
\includegraphics[width=\linewidth]{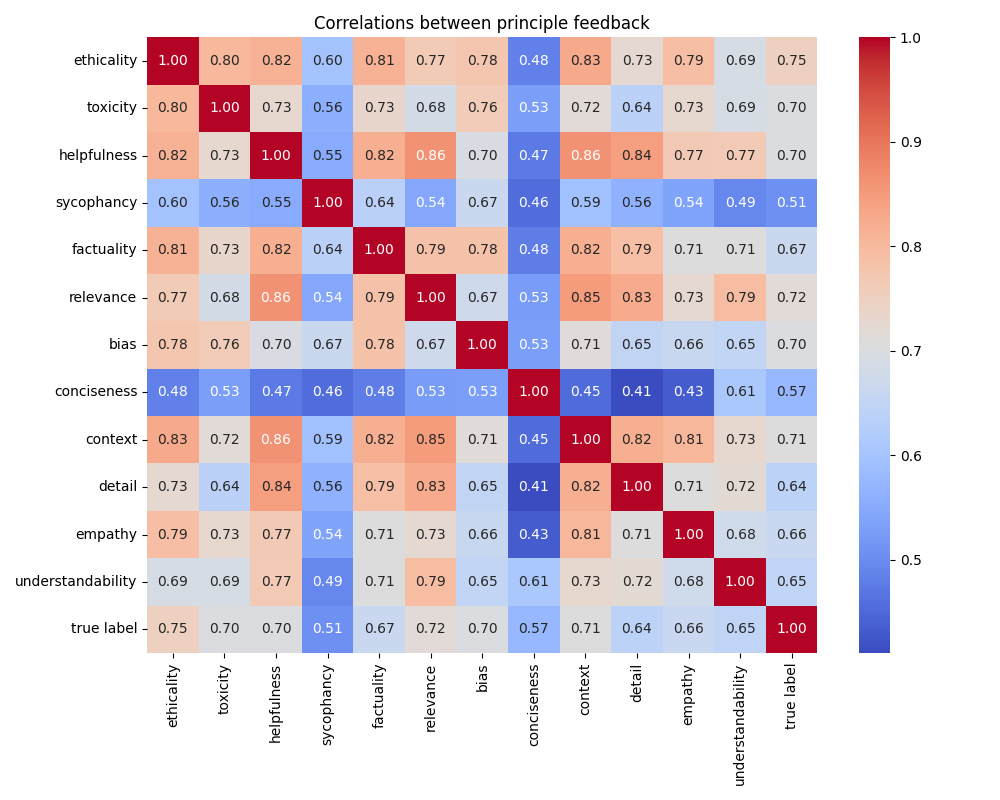}
\caption{}
\label{fig:principle_correlations}
\end{subfigure}
\hfill
\begin{subfigure}[b]{0.49\textwidth}
\centering
\includegraphics[width=\linewidth]{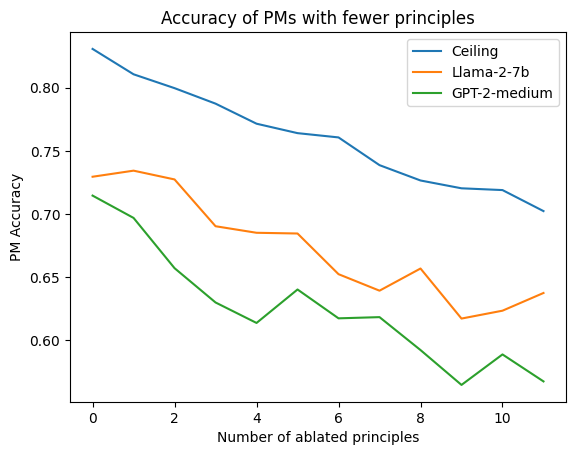}
\caption{}
\label{fig:ablation}
\end{subfigure}
\caption{(a) shows the correlations between the feedback for the different principles. (b) shows how the multi-objective PM accuracy depends on the number of principles for GPT-2-medium, Llama-2-7b, and the theoretical ceiling which represents 100\% accuracy on each individual principle.}
\label{fig:principles}
\end{figure}
Figure \ref{fig:principle_correlations} illustrates the correlations between the feedback for the different principles and single objective RLAIF. While some of the correlations are relatively strong, the preference models are clearly not being trained to detect exactly the same thing. Notably, sycophancy exhibits weak correlations with the other principles. Furthermore, sycophancy is the only preference model that receives a negative weight in the linear scalarization function logistic regression, indicating that the preferred responses are more sycophantic than the rejected ones. One benefit of MORLAIF is that we could explicitly set the sycophancy weight to zero or a positive weight if we want to reduce sycophancy without needing to go back and redo our feedback collection process. Doing so slightly reduces performance as measured by our dataset; however, it may increase performance on what we actually care about. MORLAIF might be a way to achieve a more aligned model despite having a flawed dataset.

Figure \ref{fig:ablation} shows how the performance decreases when we use fewer principles. We remove principles in the order of their weight, starting with the lowest, which was sycophancy.
\section{Related work}
Our work on MORLAIF builds upon and extends several lines of research, including Reinforcement Learning from Human Feedback (RLHF), Constitutional AI, Multi-Objective Reinforcement Learning (MORL), and task decomposition. In this section, we provide an overview of the most relevant prior work in each of these areas.

\subsection{Reinforcement Learning from Human Feedback (RLHF)}
Reinforcement learning from human feedback (RLHF) is a promising approach for aligning AI systems with human preferences. The key idea is to learn a reward function based on human feedback and use this learned reward to optimize the agent's behavior via reinforcement learning. \cite{christiano2017deep} pioneered this approach, using human preferences to train a reward model for Atari games and robotic control tasks. Subsequent work has extended RLHF to more complex domains, including summarization (\cite{stiennon2020learning,ziegler2019finetuning}), dialogue (\cite{jaques2019way}), and language modeling (\cite{ouyang2022training}).
A core component of RLHF is the reward modeling step, where human preferences are distilled into a learned reward function. Our work builds on these ideas by decomposing the reward modeling into multiple principle-specific preference models.

\subsection{Constitutional AI and RLAIF}
Constitutional AI, proposed by \cite{bai2022constitutional}, is a framework for training AI systems to adhere to a set of predefined principles or rules, analogous to a constitution. One of the methods they propose is Reinforcement learning from AI feedback, RLAIF. RLAIF works much like RLHF except that an LLM is now choosing the best response based on principles sampled from a constitution rather than utilizing humans. The authors demonstrate Constitutional AI and RLAIF on open-ended dialogue, and work on using RLAIF for summarization has been done by \cite{lee2023rlaif}.
While both Constitutional AI and our MORLAIF approach aim to constrain model behavior based on predefined principles, there are several key differences. Constitutional AI uses a single set of principles and generates both critiques and revisions in the supervised phase, while MORLAIF trains separate preference models for each moral principle. In Constitutional AI, each principle has to be quite broad and cover most of what we mean by human preferences in order to ensure convergence. In MORLAIF, by contrast, having specific and unique principles is preferable, giving us more fine-grained control and interpretability.

\subsection{Task Decomposition}
Task decomposition is a general strategy for simplifying complex tasks by breaking them down into more manageable subtasks. In the context of reinforcement learning, task decomposition has been used to improve sample efficiency and generalization (\cite{ghosh2018divide,peng2022ase}). MORLAIF can be seen as a form of task decomposition applied to reward modeling, where the complex task of representing human preferences is broken down into simpler principle-specific subtasks. This decomposition allows for more targeted data collection and potentially simplifies the learning problem for each preference model.

\subsection{Decomposing Human Preferences}
\cite{go2024compositional} proposed Compositional Preference Models (CPMs), which decompose the reward modeling task into multiple interpretable features. Rather than training a single preference model end-to-end, CPMs first define a set of human-interpretable features (e.g., helpfulness, specificity, factuality). They then use a prompted language model to assign scalar scores to each feature for a given query-response pair. Finally, a logistic regression model is trained to combine the feature scores into an overall preference judgment.
The authors show that CPMs are more robust compared to standard PMs and exhibit reduced overoptimization. Furthermore, CPMs allow for more interpretable and modular preference modeling, as the importance of each feature can be examined and the model's behavior adjusted by modifying the logistic regression weights without retraining.
MORLAIF shares the high-level motivation of CPMs in terms of decomposing reward modeling into simpler, more interpretable components. However, there are several key differences. First, MORLAIF trains separate preference models for each principle using preference comparisons, while CPMs obtain scalar scores for each feature from a single prompted LLM to train a single PM. Second, MORLAIF uses the preference model outputs as rewards in an RL framework to fine-tune the policy, while CPMs focus on the preference modeling stage and use a simple logistic regression to combine features. Nonetheless, both approaches demonstrate the benefits of decomposing preference modeling for improved interpretability and robustness in language model alignment.

\subsection{Using Multiple Preference Models}
\cite{coste2024reward} study the use of multiple preference models to improve robustness and mitigate overoptimization issues in RLHF. They use worst-case optimization (WCO) and uncertainty-weighted optimization (UWO), which are also tested in our paper. They find that using an ensemble of reward models and combining their outputs using conservative objectives practically eliminates overoptimization and improves performance compared to using a single reward model, especially under the realistic setting of noisy preference labels. In their paper, the different preference models are trained on the same objective; by contrast, we train preference models on different principles, though our method likely also mitigates overoptimization in a similar way.

\subsection{Distributional Preference Learning}
\cite{siththaranjan2024distributional} propose Distributional Preference Learning (DPL) to address hidden context in preference learning and RLHF. Hidden context, such as annotator diversity or varying annotation objectives, can lead to suboptimal reward modeling with standard preference learning. DPL estimates a distribution over utility values to account for this hidden context. Applied to an RLHF dataset, DPL identifies the hidden context of helpfulness vs. harmlessness without supervision. Optimizing a lower quantile of the DPL utility distributions also reduces jailbreak vulnerabilities caused by conflicting objectives. While MORLAIF decomposes the reward into separate models for each principle, DPL learns a single distributional model capturing uncertainty from hidden context, providing a complementary approach to improving interpretability and robustness in RLHF reward modeling.
\section{Conclusion}
In this paper, we introduced Multi-Objective Reinforcement Learning from AI Feedback (MORLAIF), a novel approach to improving the alignment and performance of language models trained using Reinforcement Learning from AI Feedback (RLAIF). By decomposing the complex task of representing human preferences into multiple principle-specific preference models, MORLAIF allows for more targeted data collection, interpretability, and control over the model's behavior.
Our experiments demonstrate that MORLAIF outperforms standard single-objective RLAIF baselines across a range of model sizes. Surprisingly, the choice of scalarization function did not significantly impact the results, suggesting that the benefits of MORLAIF may be more closely tied to the decomposition of preferences rather than the specific method of aggregation. This finding simplifies the selection of scalarization functions but may also indicate that the theoretical benefits of certain scalarization functions are limited in practice.
Furthermore, our results show that MORLAIF can be used to align larger language models using smaller preference models, which could be a promising approach for scalable oversight. However, the benefits of MORLAIF seem to diminish with increasing model scale, suggesting that its effectiveness may be reduced when applied to frontier-scale models.
Despite these limitations, MORLAIF offers several appealing properties, such as improved interpretability, the ability to adjust model behavior without retraining preference models, and potential mitigation of overoptimization issues. The decomposition of preferences into specific principles also enables more fine-grained control and analysis of the model's behavior.

\bibliographystyle{gubbins/iclr2024_conference}
\bibliography{references}

\appendix
\section{Feedback Model and Win-Rate Prompts}
\label{sec:prompts}
For feedback, the following prompt was used:
\begin{tcolorbox}[colback=purple!10!white, colframe=purple!50!black, title=Feedback Model Prompt without CoT]
You will be given a conversation between a human and an AI assistant.
Your job is to judge which of the two possible assistant responses is {principle}.
Here is the conversation: {conversation}
Here are the possible response options:
Option A: {responseA}
Option B: {responseB}
Please judge which option is {principle}, regardless of how good
the answer is from other perspectives. Please respond only with A or B.
The best response is option:
\end{tcolorbox}
\begin{tcolorbox}[colback=blue!10!white, colframe=blue!50!black, title=Feedback Model Prompt with CoT]
You will be given a conversation between a human and an AI assistant.
Your job is to judge which of the two possible assistant responses is {principle}.
Here is the conversation: {conversation}
Here are the possible response options:
Option A: {responseA}
Option B: {responseB}
Please judge which option is {principle}, regardless of how good the answer is from other perspectives. Begin your response with your reasoning inside \textless reasoning\textgreater \textless /reasoning\textgreater tags.
End your response with "Chosen option: " followed by A or B.
\end{tcolorbox}
\begin{tcolorbox}[colback=green!10!white, colframe=green!50!black, title=Win-Rate Prompt]
You will be given a conversation between a human and an AI assistant.
Your job is to judge which of the two possible assistant responses is better.
Here is the conversation: {conversation}
Here are the possible response options:
Option A: {responseA}
Option B: {responseB}
Please judge which option is better. Please respond only with A or B.
The best response is option:
\end{tcolorbox}
\section{Human Experiments}
\label{sec:human}
This section discusses some specifics of our human crowdworker win-rate experiment. Crowdworkers were recruited from the online crowdworking platform Amazon MTurk.
\begin{figure}[H]
\centering
\includegraphics[scale=0.42]{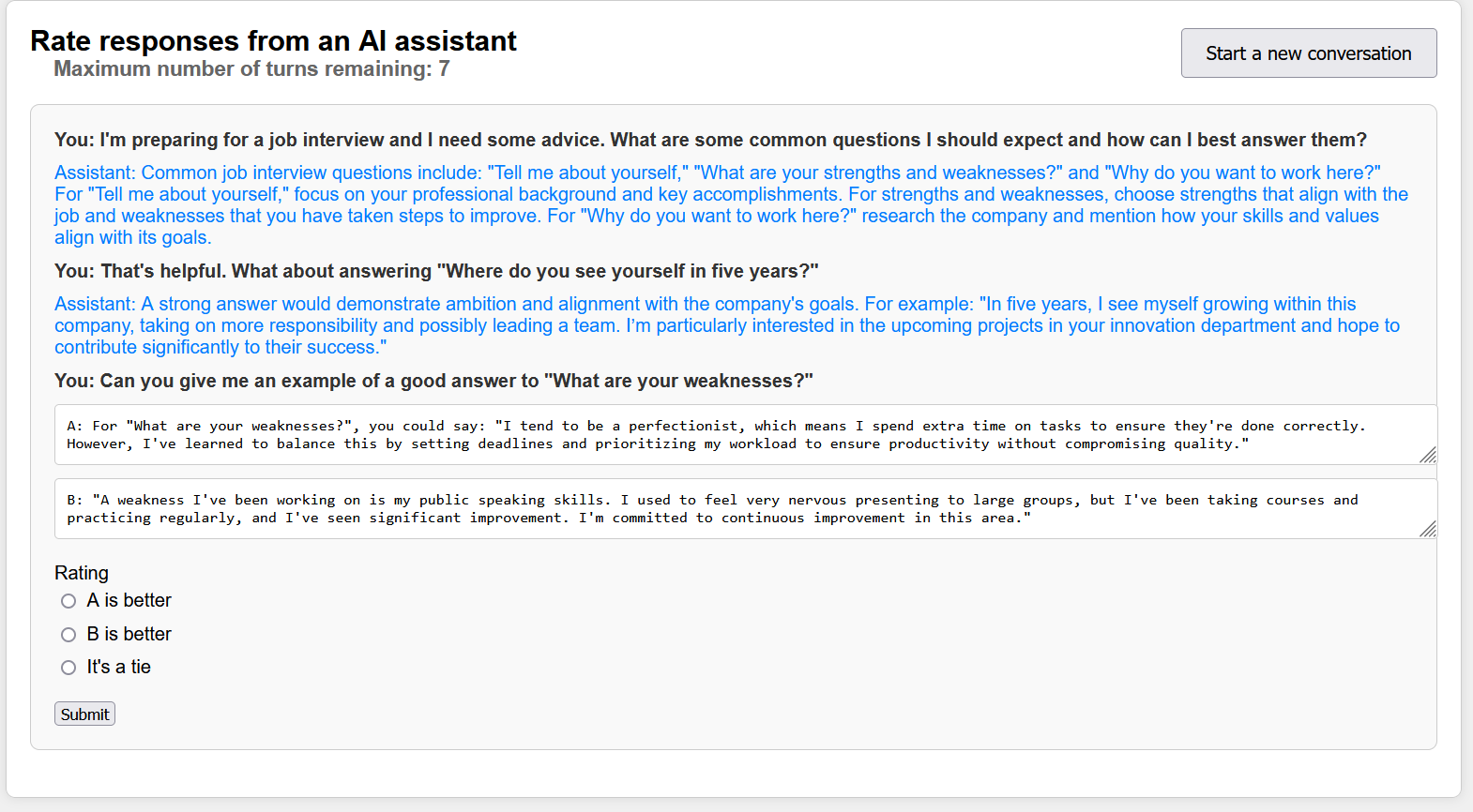}
\caption{An example showing the interface which was presented to the crowdworkers and an example conversation}
\label{fig:interface}
\end{figure}
After filling in a short form and reading the instructions, the crowdworkers were given a link which took them to a website with the interface shown in figure \ref{fig:interface}. Users were asked to have a conversation with an AI assistant. After each message, the two different models generated responses which were shown as option A and option B (in random order to mitigate positional biases). The users were then asked to rate which of the two options was better or if it was a tie. If the user chose A or B, the conversation continued with that response, while if they chose tie, a random response was added to the conversation. The conversation could continue for a maximum of 10 turns, and there was also a button to start a new conversation instead of continuing the current one. The average conversation lasted 5.2 turns. Each worker could have a maximum of 10 conversations.

We ran two win-rate experiments, one comparing a multi-objective linear Llama-2-7b with a single-objective version and one comparing a multi-objective linear Llama-2-7b trained using GPT-2-medium preference models with the same single-objective Llama-2-7b trained with a single preference model. For these comparisons, we collected 4238 and 4281 datapoints, respectively.

According to \cite{veselovsky2023prevalence}, LLM use is prevalent on crowdworking platforms. If crowdworkers use LLMs to generate their dialogue or choose their preferred responses, that invalidates the entire point of using humans in the first place. In order to mitigate this, a few steps were taken. Firstly, workers were instructed not to use LLMs and tick a box indicating that they understood this. Secondly, copy-pasting was turned off, limiting the speedup using an LLM would entail. Thirdly, before starting, users were prompted with an "LLM-proof riddle" in order to weed out people using LLMs. The task was a variation of an actual riddle where all the hard bits have been removed and there is only a trivial answer, but LLMs typically answer with the solution to the original riddle rather than the modified one. The riddle used was the following:
\begin{tcolorbox}[colback=purple!10!white, colframe=purple!50!black, title=``LLM proof'' Riddle]
There is a goat and a man on one side of a river. There's a boat
which can only carry two at a time and can only be rowed manually
by the man. How can they cross the river?
\end{tcolorbox}

LLMs very often fail to notice the trivial solution of "They both take the boat across the river" since they have been trained on the original riddle.
Finally, we checked responses with GPTZero, a tool which, despite high error rates, predicts how likely it was that a text was generated by AI. We discarded all data from users who scored above a 25\% chance of being AI, which represented about 13\% of our collected data.
\end{document}